\documentclass[10pt,twocolumn,letterpaper]{article}

\usepackage{iccv}
\usepackage{times}
\usepackage{epsfig}
\usepackage{graphicx}
\usepackage{amsmath}
\usepackage{amssymb}

\usepackage{booktabs}
\usepackage{multirow}
\usepackage{enumitem}
\usepackage{verbatim}
\usepackage{appendix}
\usepackage{subcaption}
\usepackage{makecell}

\usepackage[accsupp]{axessibility}  % Improves PDF readability for those with disabilities.

\usepackage[pagebackref=true,breaklinks=true,letterpaper=true,colorlinks,bookmarks=false]{hyperref}

\iccvfinalcopy % *** Uncomment this line for the final submission

\def\iccvPaperID{5975} % *** Enter the ICCV Paper ID here

\ificcvfinal\fi

\begin{document}

%%%%%%%%% TITLE
% \title{DDColor: Towards Photo-Realistic and Semantic-Aware Image Colorization \protect\\ via Query-based Dual Decoders}
\title{DDColor: Towards Photo-Realistic Image Colorization via Dual Decoders}

\author{
Xiaoyang Kang\quad\quad
Tao Yang\quad\quad
Wenqi Ouyang\quad\quad
Peiran Ren\quad\quad
Lingzhi Li\quad\quad
Xuansong Xie
\\
DAMO Academy, Alibaba Group
\\
{\tt\small \{kangxiaoyang.kxy,baiguan.yt,wenqi.oywq,peiran.rpr,llz273714,xingtong.xxs\}@alibaba-inc.com}
\\
}

% \and

% \maketitle

\newcommand{\methodname}{DDColor}

\twocolumn[{
\renewcommand\twocolumn[1][]{#1}
\maketitle
\vspace{-1.2cm}
\begin{center}
    \centering
    \includegraphics[width=1.0\linewidth]{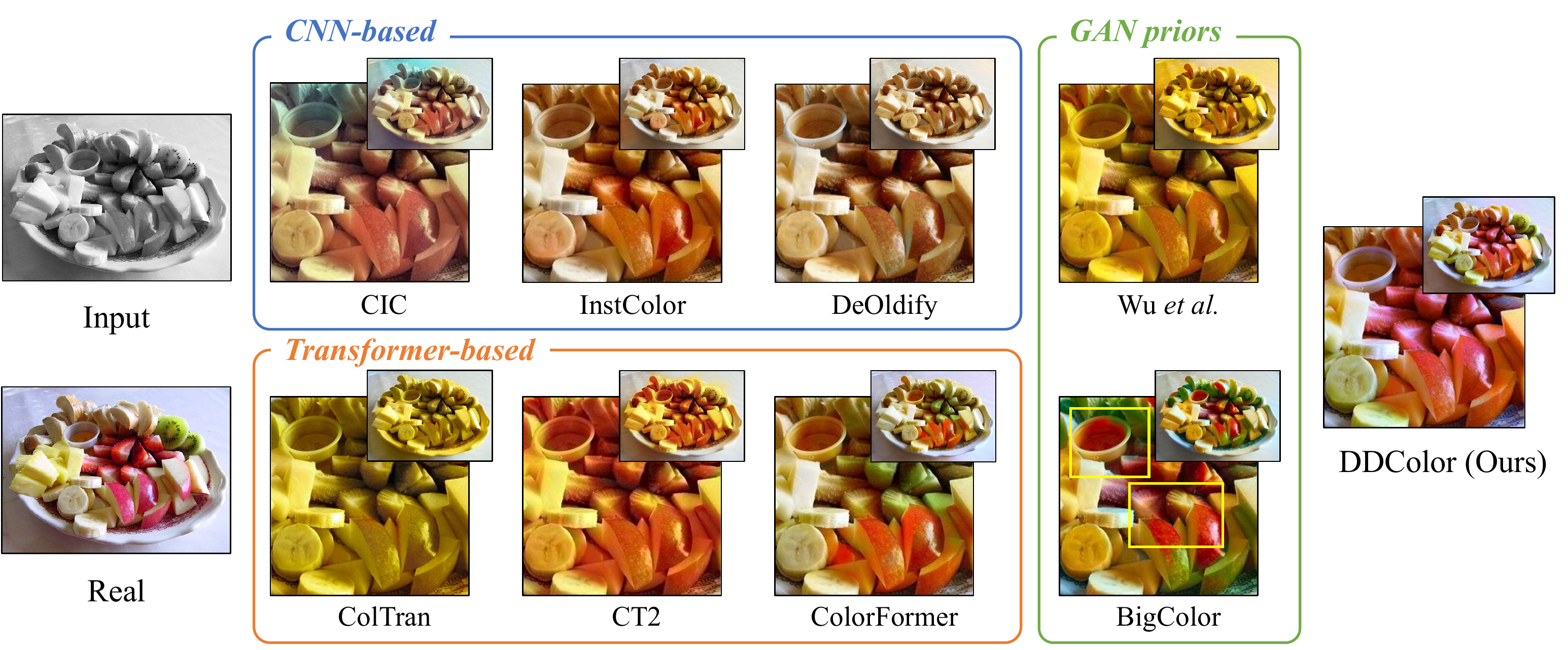}
    \captionof{figure}{\textbf{Visual comparison.} We present a novel colorization method \textit{\methodname{}}, which is capable of producing more natural and vivid colorization in complex scenes containing multiple objects with diverse contexts, compared to existing methods.}
    \label{fig:teaser}
\end{center}
\vspace{0.5cm}
}]

%%%%%%%%% ABSTRACT
\begin{abstract}
\vspace{-0.2cm}
Image colorization is a challenging problem due to multi-modal uncertainty and high ill-posedness. 
Directly training a deep neural network usually leads to incorrect semantic colors and low color richness.
While transformer-based methods can deliver better results, they often rely on manually designed priors, suffer from poor generalization ability, and introduce color bleeding effects. 
To address these issues, we propose \textbf{DDColor}, an end-to-end method with \textbf{d}ual \textbf{d}ecoders for image \textbf{color}ization. Our approach includes a pixel decoder and a query-based color decoder. The former restores the spatial resolution of the image, while the latter utilizes rich visual features to refine color queries, thus avoiding hand-crafted priors. Our two decoders work together to establish correlations between color and multi-scale semantic representations via cross-attention, significantly alleviating the color bleeding effect. Additionally, a simple yet effective colorfulness loss is introduced to enhance the color richness. Extensive experiments demonstrate that DDColor achieves superior performance to existing state-of-the-art works both quantitatively and qualitatively. 
The codes and models are publicly available at \url{https://github.com/piddnad/DDColor}.
\end{abstract}

\vspace{-0.2cm}
%%%%%%%%% BODY TEXT
\section{Introduction}
Image colorization is a classic computer vision task and has great potential in many real-world applications, such as legacy photo restoration \cite{tsaftaris2014novel}, video remastering \cite{iizuka2019deepremaster} and art creation \cite{qu2006manga}, etc. Given a grayscale image, colorization aims to recover its two missing color channels, which is highly ill-posed and usually suffers from multi-modal uncertainty, \eg, an object may have multiple plausible colors. Traditional colorization methods address this problem mainly based on user guidance such as reference images \cite{welsh2002transferring, ironi2005colorization, gupta2012image, liu2008intrinsic, chia2011semantic} and color graffiti \cite{levin2004colorization, yatziv2006fast, qu2006manga, luan2007natural}. Although great progress has been made, it remains a challenging research problem. 

With the rise of deep learning, automatic colorization has drawn a lot of attention, targeting at producing appropriate colors from complex image semantics (\eg, shape, texture, and context). Some early methods \cite{cheng2015deep, deshpande2017learning, zhang2016colorful, su2020instance, deoldify} attempt to predict per-pixel color distributions using convolutional neural networks (CNNs). Unfortunately, these CNN-based methods often yield incorrect or unsaturated colorization results due to the lack of a comprehensive understanding of image semantics (Figure~\ref{fig:teaser} CIC \cite{zhang2016colorful}, InstColor \cite{su2020instance} and DeOldify \cite{deoldify}). In order to embrace semantic information, some methods \cite{wu2021towards, Kim2022BigColor} resort to generative adversarial networks (GANs) and utilize their rich representations as generative priors for colorization. However, due to the limited representation space of GAN prior, they fail to handle images with complex structures and semantics, resulting in inappropriate colorization results or unpleasant artifacts (Figure~\ref{fig:teaser} Wu \etal \cite{wu2021towards} and BigColor \cite{Kim2022BigColor}).

With the tremendous success in natural language processing (NLP), Transformer \cite{vaswani2017attention} has been extended to many computer vision tasks. Recently, some works \cite{kumar2021colorization, Weng2022CT2, Ji2022ColorFormer} introduce the non-local attention mechanism of transformer to image colorization. Though achieving promising results, these methods either train several independent subnets, leading to accumulated error  (Figure~\ref{fig:teaser} ColTran \cite{kumar2021colorization}), or perform color attention operations on single-scale image feature maps, causing visible color bleeding when tackling complex image contexts (Figure~\ref{fig:teaser} CT2 \cite{Weng2022CT2} and ColorFormer \cite{Ji2022ColorFormer}). In addition, these methods often rely on hand-crafted dataset-level empirical distribution priors, such as color masks in \cite{Weng2022CT2} and semantic-color mappings in \cite{Ji2022ColorFormer}, which are cumbersome and difficult to generalize.

In this paper, we propose an novel colorization method, namely \methodname{}, targeting at achieving semantically reasonable and visually vivid colorization. 
Our approach utilizes an encoder-decoder structure where the encoder extracts image features and the dual decoders restore spatial resolution. 
Unlike previous methods that optimize color likelihood resorting to an extra network or manually calculated priors, our method uses a query-based transformer as color decoder to learn semantic-aware color queries in an end-to-end way.
By using multi-scale image features to learn color queries, our method alleviates color bleeding and improves the colorization of complex contexts and small objects significantly (see Figure~\ref{fig:teaser}).
Over and above this, we present a new colorfulness loss to improve the color richness of generated results.

We validate the performance of our model on public benchmarks ImageNet \cite{russakovsky2015imagenet} and conduct ablations to demonstrate the advantages of our framework. The visualization results and evaluation metrics show that our work achieves significant improvements to previous state-of-the-art methods in terms of semantic consistency, color richness, etc. Furthermore, we test our model on two additional datasets (COCO-Stuff \cite{caesar2018coco}, and ADE20k \cite{zhou2017scene}) without fine-tuning and achieve best performance among all baselines, demonstrating its generalization ability.

\begin{figure*}[t]
  \centering
  \includegraphics[width=1.0\linewidth]{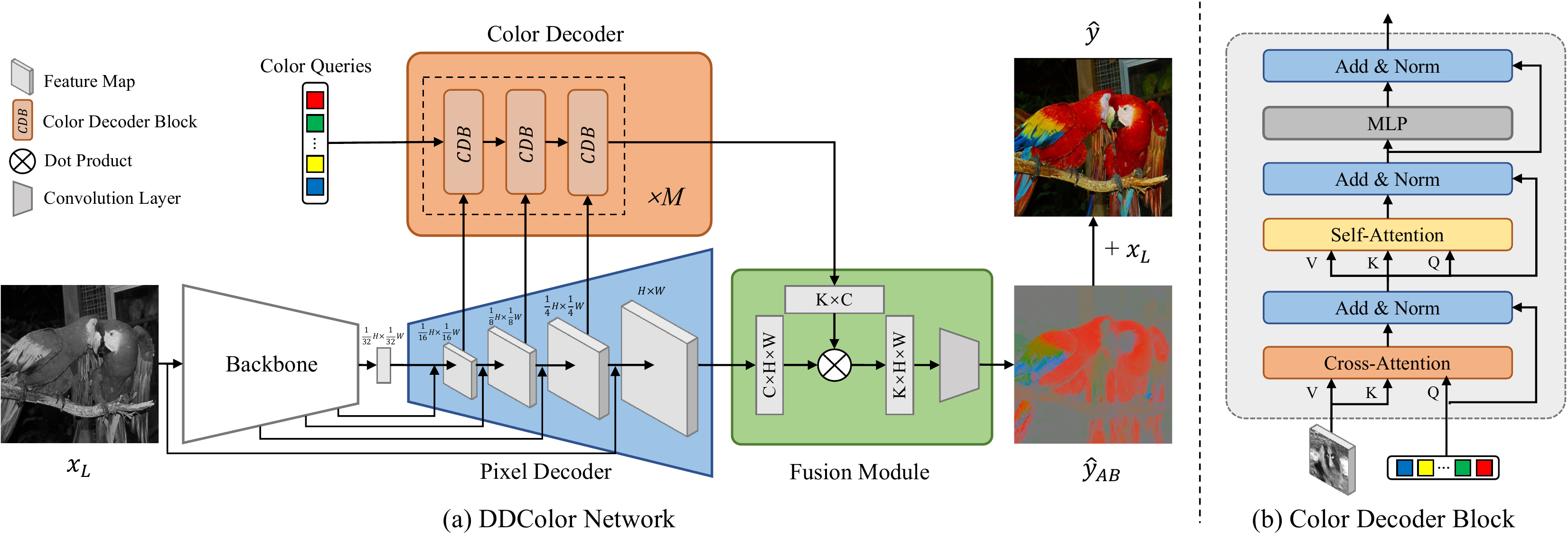}
  \caption{\textbf{(a) Method overview.} Our proposed model, DDColor, colorizes a grayscale image $x_L$ in an end-to-end fashion. We first extract its features using a backbone network, which are then input to the pixel decoder to restore the spatial structure of the image. Concurrently, the color decoder performs color queries on visual features of different scales, learning semantic-aware color representations. The fusion module combines the outputs of both decoders to produce a color channel output $\hat{y}_{AB}$. Finally, we concatenate $\hat{y}{AB}$ and $x_L$ along the channel dimension to obtain the final colorization result $\hat{y}$. \textbf{(b) Structure of color decoder block.} Taking image features and color queries as inputs, the color decoder block establishes the correlation between semantic and color representation by performing cross-attention, self-attention and feed forward operations. }
  \label{fig:network}
\end{figure*}

Our key contributions are summarized as follow:

\begin{itemize}[itemsep=0pt,topsep=0pt,parsep=0pt]
    \item We propose an end-to-end network with dual decoders for automatic image colorization, which ensures vivid and semantically consistent results.
    \item Our method includes a novel color decoder that learns color queries from visual features without relying on hand-crafted priors. Additionally, our pixel decoder provides multi-scale semantic representations to guide the optimization of color queries, which effectively reduces the color bleeding effect.
    \item Comprehensive experiments demonstrate that our method achieves state-of-the-art performance and exhibits good generalization compared to the baselines.
\end{itemize}

%-------------------------------------------------------------------------

\section{Related Work}

\noindent\textbf{Automatic colorization.} 
The emergence of large-scale datasets and the development of DNNs make it possible to colorize grayscale images in a data-driven manner. Cheng \etal \cite{cheng2015deep} propose the first DNN-based image colorization method. Zhang \etal \cite{zhang2016colorful} learn the color distribution of each pixel and train the network with a rebalanced multinomial cross-entropy, allowing rare colors to appear. MDN \cite{deshpande2017learning} uses a variational autoencoder (VAE) to get diverse colorized results. InstColor \cite{su2020instance} believes that a clear figure-ground separation helps to improve performance of colorization, thus adopts a detection model to provide the detection box as prior. Later works \cite{zhao2018pixel, zhao2020pixelated} extend to take segmentation mask as pixel-level object semantics to guide colorization. Recently, some works \cite{wu2021towards, Kim2022BigColor} attempt to restore vivid color by taking advantage of the rich and diverse color priors of pre-trained GANs.

\noindent\textbf{Vision transformer for colorization.} 
Since successfully introduced Transformer \cite{vaswani2017attention} to vision recognition, Vision Transformer (ViT) \cite{dosovitskiy2020image} has developed rapidly in many downstream vision tasks \cite{carion2020end, zhu2020deformable, zheng2021rethinking, cheng2021per}. In the field of colorization, ColTran \cite{kumar2021colorization} first uses a transformer to build a probability model, and samples color from the learning distribution to conditionally generate a low-resolution coarse colorization, before upsampling it into a high-resolution image of fine color. CT2 \cite{Weng2022CT2} considers colorization as a classification task, and feeds image patch and color tokens together into a ViT-based network including a luminance-selecting module with pre-calculated probability distribution of dataset. ColorFormer \cite{Ji2022ColorFormer} proposes a transformer network with hybrid self-attention, and refines image features by a memory network with pre-built semantic-color priors. In this work, we introduce a color decoder that enables end-to-end learning of color queries from multi-scale visual features, eliminating the need for hand-crafted priors.

\noindent\textbf{Query-based transformer in computer vision.} 
Recently, many researchers have been using query-based transformer for various tasks due to its ability to leverage attention mechanisms to capture global correlations. DETR \cite{carion2020end} is the first to introduce transformers to object detection, using queries to locate and represent candidate objects. Following DETR, MaskFormer \cite{cheng2021per} and QueryInst \cite{fang2021instances} respectively introduce query-based transformers to semantic and instance segmentation, showing its great potential to vision tasks. Transtrack \cite{sun2020transtrack} applies queries across the frames to improve multi-object tracking. In this work, we apply query-based transformer for image colorization for the first time.

%-------------------------------------------------------------------------

\section{Method}
\subsection{Overview}

Given a grayscale input image $x_L \in \mathbb{R}^{H \times W \times 1}$, our colorization network predicts the two missing color channels $\hat{y}_{AB} \in \mathbb{R}^{H \times W \times 2}$, where the $L$, $AB$ channels represent the luminance and chrominance in CIELAB color space, respectively. The network adopts an encoder-decoder framework, as shown in Figure~\ref{fig:network} (a). 

We utilize a backbone network as the encoder to extract high-level semantic information from grayscale images.
The backbone network is designed to extract image semantic embedding, which is crucial for colorization. In this work, we choose ConvNeXt \cite{liu2022convnet}, which is the cutting-edge model for image classification. Taken $x_L$ as input, the backbone network outputs $4$ intermediate feature maps with resolutions of $\frac{H}{4}\times\frac{W}{4} $, $\frac{H}{8}\times\frac{W}{8}$, $\frac{H}{16}\times\frac{W}{16}$ and $\frac{H}{32}\times\frac{W}{32}$. The first three feature maps are fed to pixel decoder through shortcut connections, while the last is treated as input to the pixel decoder. 
As for the backbone network structure, there are several options, such as ResNet\cite{he2016deep}, Swin-Transformer\cite{liu2021swin}, etc., as long as the network is capable of producing a hierarchical representation.

The decoder section of our framework consists of a pixel decoder and a color decoder.
The pixel decoder uses a series of stacked upsampling layers to restore the spatial resolution of the image features. Each upsampling layer has a shortcut connection with the corresponding stage of the encoder.
The color decoder gradually refines semantic-aware color queries by leveraging multiple image features at different scales.
Finally, the image and color features produced by the two decoders are fused to generate the color output. 

In the following, we provide detailed descriptions of these modules as well as the losses used for colorization.

\subsection{Dual Decoders}

\subsubsection{Pixel Decoder}

The pixel decoder is composed of four stages that gradually expand the image resolution. Each stage includes an upsampling layer and a shortcut layer.
Specifically, unlike previous methods that use deconvolution \cite{noh2015learning} or interpolation \cite{long2015fully}, we employ PixelShuffle \cite{shi2016real} as the upsampling layer. This layer rearranges low-resolution feature maps with the shape of $(\frac{h}{p}, \frac{w}{p}, cp^2)$ into high-resolution ones with the shape of $(h, w, c)$. The shortcut layer uses a convolution to integrate features from the corresponding stages of the encoder through shortcut connections.

Our method captures a complete image feature pyramid through a step-by-step upsampling process, which is beyond the capability of some transformer-based approaches \cite{kumar2021colorization, Weng2022CT2}. These multi-scale features are further utilized as input to the color decoder to guide the optimization of color queries. The final output of the pixel decoder is the image embedding $E_i \in \mathbb{R}^{C \times H \times W}$, which has the same spatial resolution as the input image.

\subsubsection{Color Decoder}

Many existing colorization methods rely on additional priors to achieve vivid results. For example, some methods \cite{wu2021towards, Kim2022BigColor} utilize generative prior from pretrained GANs, while others use empirical distribution statistics \cite{Weng2022CT2} or pre-built semantic-color pairs \cite{Ji2022ColorFormer} of training sets. However, these approaches require extensive pre-construction efforts and may have limited applicability in various scenarios. 
To reduce reliance on manually designed priors, we propose a novel query-based color decoder.

\noindent\textbf{Color decoder block.} The color decoder is composed of a stack of blocks, with each block receiving visual features and color queries as input.
The color decoder block (CDB) is designed based on a modified transformer decoder, as depicted in Figure~\ref{fig:network} (b).

To learn a set of adaptive color queries based on visual semantic information, we create learnable color embedding memories to store the sequence of color representations: $\mathcal{Z}_0 = [\mathcal{Z}_0^1,\mathcal{Z}_0^2,\dots,\mathcal{Z}_0^K] \in \mathbb{R}^{K \times C}$. These color embeddings are initialized to zero during the training phase and used as color queries in the first CDB. We first establish the correlation between semantic representation and color embedding through the cross-attention layer:
\begin{equation}
  \mathcal{Z}_l^{\prime} = {softmax} (Q_l K_l^T) V_l + \mathcal{Z}_{l-1},
  \label{eq:cross-attention}
\end{equation}
where $l$ is the layer index, $\mathcal{Z}_l \in \mathbb{R}^{K \times C}$ refers to $K$ $C$-dimensional color embeddings at the $l^{th}$ layer. $Q_l = f_Q(\mathcal{Z}_{l-1}) \in \mathbb{R}^{K \times C}$, and $K_l, V_l \in \mathbb{R}^{H_l \times W_l \times C}$ are the image features under the transformations $f_K(\cdot)$ and $f_V(\cdot)$, respectively. $H_l$ and $W_l$ are the spatial resolutions of image features, and $f_Q$, $f_K$ and $f_V$ are linear transformations.

With the aforementioned cross-attention operation, the color embedding representation is enriched by the image features. We then utilize standard transformer layers to transform the color embedding, as follows:
\begin{align}
  \mathcal{Z}_l^{\prime\prime} &= MSA(LN(\mathcal{Z}_{l}^{\prime})) + \mathcal{Z}_{l}^{\prime}, \label{eq:msa}\\
  \mathcal{Z}_l^{\prime\prime\prime} &= MLP(LN(\mathcal{Z}_l^{\prime\prime})) +\mathcal{Z}_l^{\prime\prime}, \label{eq:mlp}\\
  \mathcal{Z}_l &= LN(\mathcal{Z}_l^{\prime\prime\prime}), \label{eq:ln}
\end{align}
where ${MSA}(\cdot)$ indicates the multi-head self-attention\cite{vaswani2017attention}, $MLP(\cdot)$ denotes the feed forward network, and $LN(\cdot)$ is the layer normalization\cite{ba2016layer}. 
It is worth mentioning that cross-attention is operated before self-attention in the proposed CDB. This is because the color queries are zero-initialized and semantically independent before the first self-attention layer is applied. 

\noindent\textbf{Extending to multi-scale.} Previous transformer-based colorization methods often performed color attention on single-scale image feature maps and failed to adequately capture low-level semantic cues, potentially leading to color bleeding when dealing with complex contexts. In contrast, multi-scale features have been widely explored in many computer vision tasks such as object detection \cite{liu2016ssd} and instance segmentation \cite{he2017mask}. These features can boost the performance of colorization as well(see ablations in Sec~\ref{sec:ablation}).

To balance computational complexity and representation capacity, we select image features of three different scales. Specifically, we use the intermediate visual features generated by the pixel decoder with downsample rates of $1/16$, $1/8$, and $1/4$ in the color decoder. We group blocks with $3$ CDBs per group, and in each group, the multi-scale features are fed to CDBs in a sequence. We repeat the group for $M$ times in a round-robin fashion. In total, the color decoder consists of $3M$ CDBs. We can formulate the color decoder as follows:
\begin{equation}
  E_c = ColorDecoder(\mathcal{Z}_0, \mathcal{F}_1, \mathcal{F}_2, \mathcal{F}_3),
  \label{eq:color-decoder}
\end{equation}
where $\mathcal{F}_1, \mathcal{F}_2$ and $\mathcal{F}_3$ are visual features at three different scales. 

The use of multi-scale features in the color decoders can model the relationship between color queries and visual embeddings, making the color embedding $E_c \in \mathbb{R}^{K \times C}$ more sensitive to semantic information, further enabling more accurate identification of semantic boundaries and less color bleeding.

\begin{table*}[t]\footnotesize
\centering
\setlength{\tabcolsep}{0.42\tabcolsep}

\begin{tabular}{@{}lccccccccccccccccc@{}}
\toprule
\multirow{2}{*}{Method} & \multirow{2}{*}{\#Params.} & \multicolumn{4}{c}{ImageNet (val5k)} & \multicolumn{4}{c}{ImageNet (val50k)} & \multicolumn{4}{c}{COCO-Stuff}     & \multicolumn{4}{c}{ADE20K*}         \\ \cmidrule(lr){3-6}\cmidrule(lr){7-10}\cmidrule(lr){11-14}\cmidrule(lr){15-18}
 &  & FID$\downarrow$    & CF$\uparrow$     & $\Delta$CF$\downarrow$   & PSNR$\uparrow$   & FID$\downarrow$      & CF$\uparrow$     & $\Delta$CF$\downarrow$   & PSNR$\uparrow$   & FID$\downarrow$    & CF$\uparrow$    & $\Delta$CF$\downarrow$  & PSNR$\uparrow$  & FID$\downarrow$    & CF$\uparrow$    & $\Delta$CF$\downarrow$  & PSNR$\uparrow$  \\ \midrule
CIC\cite{zhang2016colorful}                     & 32.2M & 8.72  & 31.60  & 6.61        & 22.64  & 19.17   & \textbf{43.92}  & 4.83        & 20.86  & 27.88 & 33.84 & 4.40       & 22.73 & 15.31 & 31.92 & 3.12       & 23.14 \\
InstColor\cite{su2020instance}              & 69.4M & 8.06  & 24.87  & 13.34       & 23.28  & 7.36    & 27.05  & 12.04       & 22.91  & 13.09 & 27.45 & 10.79      & 23.38 & 15.44 & 23.54 & 11.50       & 24.27 \\
DeOldify\cite{deoldify}                & 63.6M  & 6.59  & 21.29  & 16.92       & \textbf{24.11}  & 3.87    & 22.83  & 16.26       & 22.97  & 13.86 & 24.99 & 13.25      & \textbf{24.19} & 12.41 & 17.98 & 17.06      & \textbf{24.40}  \\
Wu \etal\cite{wu2021towards} & 310.9M & 5.95  & 32.98  & 5.23        & 21.68  & 3.62    & 35.13  & 3.96        & 21.81  & -     & -     & -          & -     & 13.27 & 27.57 & 7.47       & 22.03 \\
ColTran \cite{kumar2021colorization}                 & 74.0M  & 6.44  & 34.50  & 3.71        & 20.95  & 6.14    & 35.50  & 3.59        & 22.30  & 14.94 & 36.27 & 1.97       & 21.72 & 12.03 & 34.58 & 0.46       & 21.86 \\
CT2 \cite{Weng2022CT2}                     & 463.0M  & 5.51  & 38.48  & 0.27        & 23.50  & 4.95       & 39.96      & 0.87           & 22.93      & -     & -     & -          & -     & 11.42 & \textbf{35.95} & 0.91       & 23.90  \\
BigColor \cite{Kim2022BigColor}                & 105.2M  & 5.36     & \textbf{39.74}      & 1.53           & 21.24      & 1.24    & 40.01  & 0.92        & 21.24  & -     & -     & -          & -     & 11.23 & 35.85 & 0.81       & 21.33 \\
ColorFormer \cite{Ji2022ColorFormer}             & 44.8M & 4.91     & 38.00      & 0.21           & 23.10      & 1.71    & 39.76  & 0.67        & 23.00  & 8.68  & 36.34 & 1.90       & 23.91 & 8.83  & 32.27 & 2.77       & 23.97 \\ \midrule
DDColor-tiny                     & 55.0M  & 4.38  & 37.66  & 0.55 & 23.54      & 1.23    & 37.72  & 1.37   & 23.63       & 7.24  & \textbf{38.48} & \textbf{0.24}   & 23.45     & 10.03  & 35.27  & \textbf{0.23} & 24.39     \\ 
DDColor-large                  & 227.9M  & \textbf{3.92}  & 38.26  & \textbf{0.05}        & 23.85  & \textbf{0.96}    & 38.65  & \textbf{0.44}        & \textbf{23.74}  & \textbf{5.18}  & \textbf{38.48} & \textbf{0.24}       & 22.85 & \textbf{8.21}  & 34.80  & \textbf{0.24}       & 24.13 \\ \bottomrule
\end{tabular}

\caption{\textbf{Quantitative comparison of different methods on benchmark datasets.} $\uparrow$ ($\downarrow$) indicates higher (lower) is better. - means the results are unavailable. Particularly, the results on ADE20K dataset are reported by running their official codes.}
\label{tab:comparison}
\end{table*}

\subsection{Fusion Module}
The fusion module is a lightweight module that combines the outputs of the pixel decoder and the color decoder to generate a color result. As shown in Figure~\ref{fig:network}, the inputs to the fusion module are the per-pixel image embedding $E_i \in \mathbb{R}^{C \times H \times W}$ from the pixel decoder, where $C$ is the embedding dimension, and the semantic-aware color embedding $E_c \in \mathbb{R}^{K \times C}$ from the color decoder, where $K$ is the number of color queries.

The fusion module aggregates these two embeddings to form an enhanced feature $\hat{\mathcal{F}} \in \mathbb{R}^{K\times H\times W}$ using a simple dot product. A $1 \times 1$ convolution layer is then applied to generate the final output $\hat{y}_{AB} \in \mathbb{R}^{2\times H\times W}$, which represents the $AB$ color channel:
\begin{align}
  \hat{\mathcal{F}} &= E_c \cdot E_i, \label{eq:fuse}\\
  \hat{y}_{AB} &= Conv(\hat{\mathcal{F}}). \label{eq:1x1conv}
\end{align}

Finally, the colorization result $\hat{y}$ is obtained by concatenating the output $\hat{y}_{AB}$ with the grayscale input $x_L$.

\subsection{Objectives}

During the training phase, the following four losses are adopted:

\noindent\textbf{Pixel loss.} The pixel loss $\mathcal{L}_{pix}$ is the L1 distance between the colorized image $\hat{y}$ and the ground truth image $y$, which provides pixel-level supervision and encourages the generator to produce outputs that are similar to the real image.

\noindent\textbf{Perceptual loss.} To ensure that the generated image $\hat{y}$ is semantically reasonable, we use a perceptual loss $\mathcal{L}_{per}$ to minimize the semantic difference between it and the real image $y$. This is accomplished using a pre-trained VGG16 \cite{simonyan2014very} to extract features from both images.

\noindent\textbf{Adversarial loss.} A PatchGAN\cite{isola2017image} discriminator is added to tell apart predicted results and real images, pushing the generator to generate indistinguishable images. Let $\mathcal{L}_{adv}$ denote the adversarial loss.

\noindent\textbf{Colorfulness loss.} We introduce a new colorfulness loss $\mathcal{L}_{col}$, inspired by the colorfulness score\cite{hasler2003measuring}. This loss encourages the model to generate more colorful and visually pleasing images. It formulates as follow:
\begin{equation}
  \mathcal{L}_{col}= 1 - [\sigma_{rgyb}(\hat{y})+0.3\cdot\mu_{rgyb}(\hat{y})] / 100,
  \label{eq:color-loss}
\end{equation}
where $\sigma_{rgyb}(\cdot)$ and $\mu_{rgyb}(\cdot)$ denote the standard deviation and mean value, respectively, of the pixel cloud in the color plane, as described in \cite{hasler2003measuring}.

The full objective for the generator is formed as follow:
\begin{equation}
  \mathcal{L}_\theta=\lambda_{pix} \mathcal{L}_{pix}+\lambda_{per} \mathcal{L}_{per}+\lambda_{adv} \mathcal{L}_{adv}+\lambda_{col} \mathcal{L}_{col},
  \label{eq:total-loss}
\end{equation}
where $\lambda_{pix}$, $\lambda_{per}$, $\lambda_{adv}$ and $\lambda_{col}$ are balancing weights of different terms.

%------------------------------------------------------------------------
\section{Experiments}

\begin{figure*}[t]
  \centering
  \includegraphics[width=1.0\linewidth]{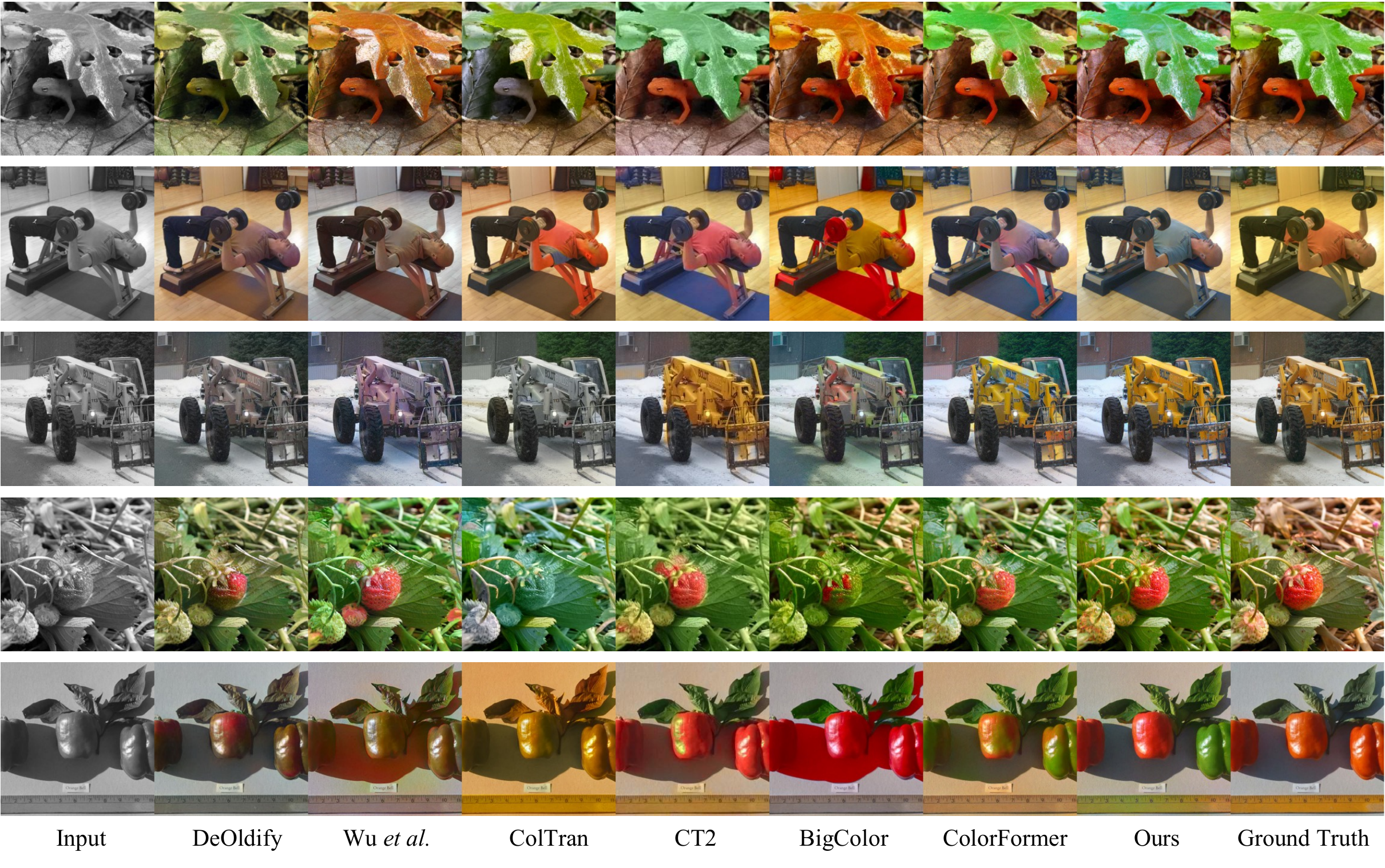}
  \caption{\textbf{Visual comparison of competing methods on automatic image colorization.} Test images are from the ImageNet validation set. One can see that our method generates more natural and vivid colors than SOTAs. Zoom in for best view.}
  \label{fig:cmp-sota}
\end{figure*}

\subsection{Experimental Setting}
\label{sec:exp-setting}
\noindent\textbf{Datasets.}
We conduct experiments on three datasets.

\textit{ImageNet} \cite{russakovsky2015imagenet} has been widely used by most existing colorization methods. It consists of 1.3M (50,000) images for training (testing). It is worthy to note that some works \cite{ardizzone2019guided, kumar2021colorization, Weng2022CT2} only use the first 5,000 images for validation. 

\textit{COCO-Stuff} \cite{caesar2018coco} contains a wide variety of natural images. We test on the 5,000 images of the original validation set without fine-tuning.

\textit{ADE20K} \cite{zhou2017scene} is composed of scene-centric images with large diversity. We test on the 2,000 images of validation set without fine-tuning.

\noindent\textbf{Evaluation metrics.}
Following the experimental protocol of existing colorization methods, we mainly use Fréchet inception distance (FID) \cite{heusel2017gans} and colorfulness score (CF) \cite{hasler2003measuring} to evaluate the performance of our method, where FID measures the distribution similarity between generated images and ground truth images and CF reflects the vividness of generated images. We also provide Peak Signal-to-Noise Ratio (PSNR) \cite{huynh2008scope} for reference, although it is a widely held view that the pixel-level metrics may not well reflect the actual colorization performance \cite{cao2017unsupervised, he2018deep, messaoud2018structural, vitoria2020chromagan, zhao2020pixelated, su2020instance, wu2021towards, Ji2022ColorFormer}.

\noindent\textbf{Implementation details.}
We train our network with AdamW \cite{loshchilov2017decoupled} optimizer and set $\beta_1 = 0.9$, $\beta_2 = 0.99$, weight decay = 0.01. The learning rate is initialized to $1e^{-4}$. For the loss terms, we set $\lambda_{pix}$ = 0.1, $\lambda_{per}$ = 5.0, $\lambda_{adv}$ = 1.0 and $\lambda_{col}$ = 0.5. 
We use ConvNeXt-L as the backbone network. For the pixel decoder, the feature dimensions after four upsampling stages are 512, 512, 256, and 256, respectively. For the color decoder, we set $M$ = 3, $K$ = 100. 
The whole network is trained in an end-to-end self-supervised fashion for 400,000 iterations with batch size of 16 and the learning rate is decayed by 0.5 at 80,000 iterations and every 40,000 iterations thereafter. We adopt color augmentation\cite{Kim2022BigColor} to real color images during training. The training images are resized into 256 $\times$ 256 resolution. All experiments are conducted on 4 Tesla V100 GPUs.

\subsection{Comparison with State-of-the-Art Methods}
\label{sec:cmp-sota}

\noindent\textbf{Quantitative comparison.} 
We benchmark our method against previous methods on three datasets and report quantitative results in Table~\ref{tab:comparison}. For all previous methods, we conducted tests using their official codes and weights.
On the ImageNet dataset, our method achieves the lowest FID, indicating that it can produce high-quality and high-fidelity colorization results. 
In particular, when the model size is comparable, our method still outperforms previous state-of-the-arts, \eg, ColorFormer \cite{Ji2022ColorFormer}.
Our method also achieves the lowest FID on the COCO-Stuff and ADE20K datasets, which demonstrates the generalization ability of our method.
The colorfulness score can reflect the vividness of the image. It can be seen that some methods \cite{zhang2016colorful, Weng2022CT2, Kim2022BigColor} report higher scores than ours. However, high colorfulness score does not always mean good visual quality (see the 6th column of Figure~\ref{fig:cmp-sota}). Therefore, we further calculate $\Delta$CF to report the colorfulness score difference between the generated image and the ground truth image. Our method achieves the lowest $\Delta$CF on all datasets, indicating that our method achieves more natural and realistic colorization.

\begin{table*}[t!]
\label{tab:ablation}
    \begin{subtable}[t]{0.3\textwidth}
        \centering
        \resizebox{\textwidth}{!}{
        \begin{tabular}{ccccc}
          \toprule
          ColorDec. & $\mathcal{L}_{col}$ & FID$\downarrow$ & CF$\uparrow$ & $\Delta$CF$\downarrow$ \\
          \midrule
          $\times$ & $\times$ & 6.04 & 33.07 & 5.14 \\
          $\times$ & \checkmark & 5.93 & 36.14 & 2.07 \\
          \checkmark & $\times$ & 4.01 & 35.69 & 2.52 \\
          \checkmark & \checkmark & \textbf{3.92} & \textbf{38.26} & \textbf{0.05} \\
          \bottomrule
        \end{tabular}
        }
        \caption{Color Decoder and Colorfulness Loss.}
        \label{tab:ablation_module}
    \end{subtable}
    \hfill
    \begin{subtable}[t]{0.325\textwidth}
        \centering
        \resizebox{\textwidth}{!}{
            \begin{tabular}{lccc}
              \toprule
              Feature Scales & FID$\downarrow$ & CF$\uparrow$ & $\Delta$CF$\downarrow$ \\
              \midrule
              single scale (1/16) & 5.09 & 37.22 & 0.99 \\
              single scale (1/8) & 4.49 & 37.58  & 0.63 \\
              single scale (1/4) & 4.44 & 37.74 & 0.47 \\
              multi-scale (3 scales) & \textbf{3.92} & \textbf{38.26} & \textbf{0.05} \\
              \bottomrule
            \end{tabular}
        }
        \caption{Different Feature Scales.}
        \label{tab:ablation_scale}
    \end{subtable}
    \hfill
    \begin{subtable}[t]{0.325\textwidth}
        \centering
        \resizebox{\textwidth}{!}{
            \begin{tabular}{lccc}
              \toprule
              Decoder Architecture & FID$\downarrow$ & CF$\uparrow$ & $\Delta$CF$\downarrow$ \\
              \midrule
              self-attn. + self-attn.  & 8.74 & 51.98 & 13.77\\
              cross-attn. + cross-attn. & 4.55 & 39.93 & 1.72 \\
              self-attn. + cross-attn. & 3.98 & 37.70 & 0.51 \\
              cross-attn. + self-attn. & \textbf{3.92} & \textbf{38.26} & \textbf{0.05} \\
              \bottomrule
            \end{tabular}
        }
        \caption{Color Decoder Architecture.}
        \label{tab:ablation_decoder}
    \end{subtable}
    \caption{\textbf{Ablation studies.} All the experiment are conducted using ImageNet (val5k) validation set.}
\end{table*}

\noindent\textbf{Qualitative comparison.}
We visualize the image colorization results in Figure~\ref{fig:cmp-sota}. 
Note that ground truth images are for reference only, and the evaluation criteria should not be color similarity due to the multi-modal uncertainty of the problem. 
It is observable that our results are more natural, more vivid, and suffer less from the color bleeding compared with other competitors.
As we can see, DeOldify \cite{deoldify} tends to produce dull and unsaturated images. 
ColTran \cite{kumar2021colorization} accumulates errors because the three subnets are trained independently, leading to noticeable unnatural colorization results, such as lizards (row 1) and vegetables (row 5).
Wu \etal \cite{wu2021towards} and BigColor \cite{Kim2022BigColor}, both based on the GAN prior, produce unpleasant red artifacts on shadows (row 2 and 5) and vehicles (row 3). 
CT2 \cite{Weng2022CT2} and ColorFormer \cite{Ji2022ColorFormer} occasionally produce incorrect colorization results especially in scenarios with complex image semantics (the person in row 2). Additionally, visible color bleeding effects can also be observed in the results (the vehicle in row 3, the strawberry in row 4 and vegetables in row 5).
Instead, our approach generates semantically reasonable and visually pleasing colorization results for complex scenes such as lizards and leaves (row 1) or the person in the gym (row 2), and successfully maintains the consistent tone and captures the details of salient object in a picture such as vehicles (row 3) and strawberries (row 4).
Interestingly, it also produces a variety of colors for objects, such as chili peppers (row 5). We attribute this to colorfulness loss, which encourages the model to produce more vivid results and better align with human aesthetics. More results can be found in the supplementary material.

\begin{figure}[ht]
  \centering
  \includegraphics[width=0.95\linewidth]{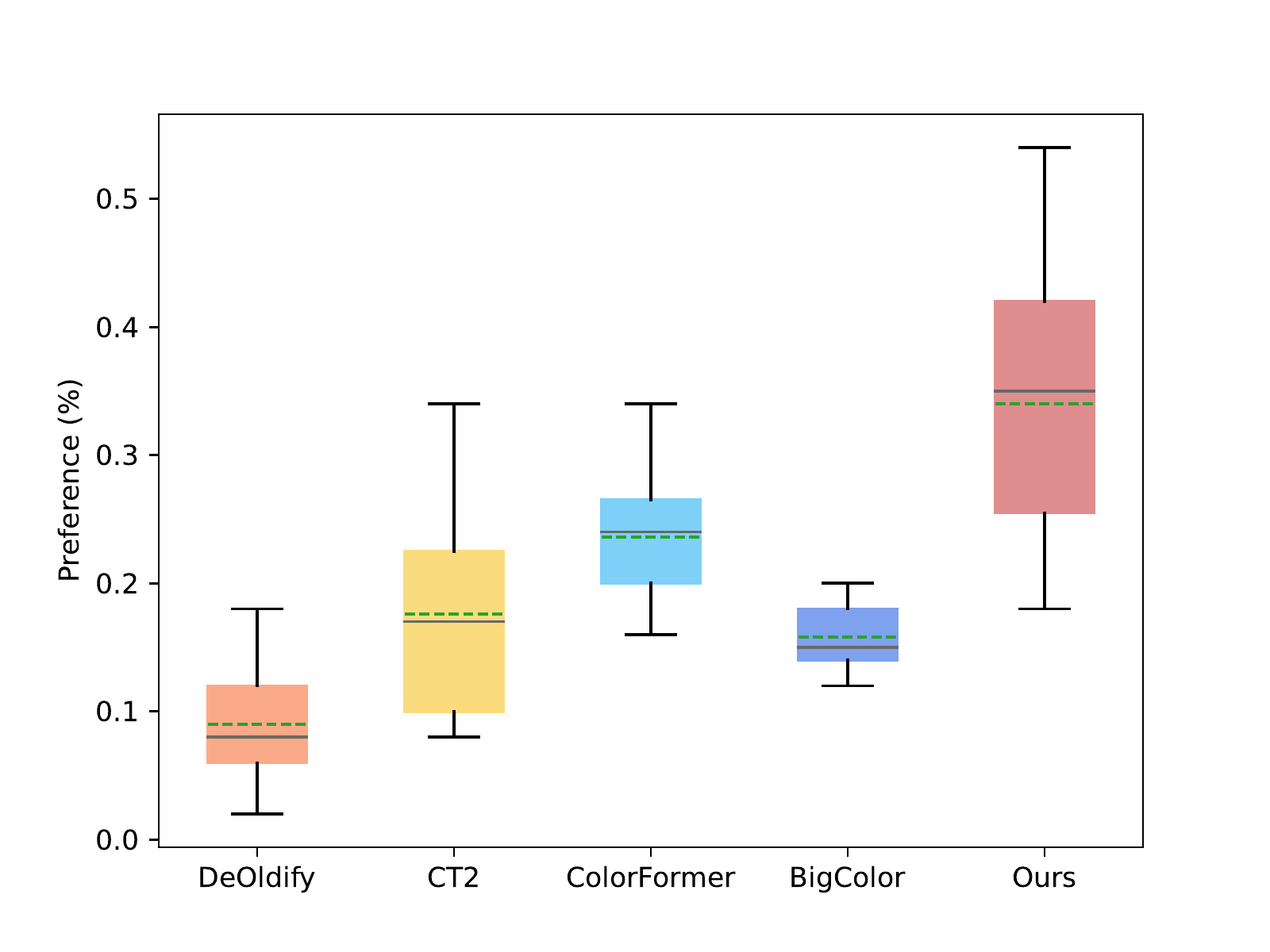}
  \caption{\textbf{Boxplot of user study.} The dashed green line and the solid gray line inside the bars are the mean and the median preference percentage, respectively.}
  \label{fig:boxplot}
\end{figure}

\noindent\textbf{User study.} We conduct a user study to investigate the subjective preference of human observers for each colorization method. Specifically, we compare our method with DeOldify\cite{deoldify}, BigColor\cite{Kim2022BigColor}, CT2\cite{Weng2022CT2} and ColorFormer\cite{Ji2022ColorFormer}. We randomly select 50 input images from the ImageNet validation set together with the coloring results displayed to 20 subjects. Subjects select the best colorized image from the randomly shuffle results of different methods. As shown in Figure~\ref{fig:boxplot}, our method is preferred by a wider range of users than the state-of-the-art methods.

\subsection{Ablation Study}
\label{sec:ablation}

\begin{figure}[t]
  \centering
  \includegraphics[width=1.0\linewidth]{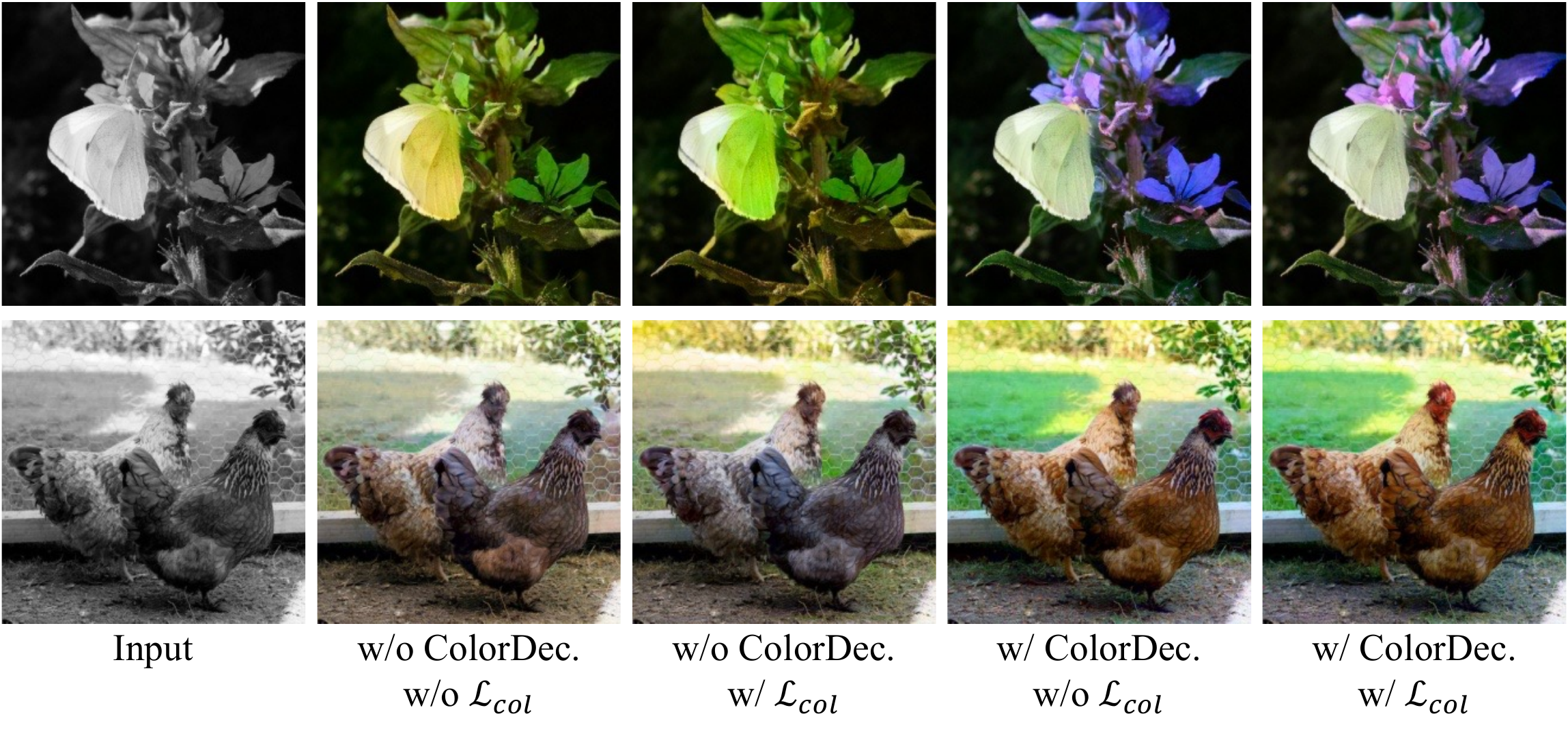}
  \caption{\textbf{Visual results} of ablation on modules.}
  \label{fig:ablation-module}
\end{figure}

\noindent\textbf{Color decoder and colorfulness loss.}
We construct a variant of our model that excludes color decoder, \ie, the entire network structure contains only the backbone and the pixel decoder. We then train both our full model and its variant twice, with and without colorfulness loss. As shown in Table~\ref{tab:ablation_module} and Figure~\ref{fig:ablation-module}, the proposed color decoder plays an important role in the final colorization result because of the adaptive color queries learned from diverse semantic features. Compared with baselines, the method with color decoder can achieve more natural and semantically reasonable colorization on diverse objects (such as butterflies and flowers in row 1, roosters and lawns in row 2). It can also be seen that the introducing colorfulness loss helps improve the colorfulness of the final result. 

\begin{figure}[t]
  \centering
  \includegraphics[width=1.0\linewidth]{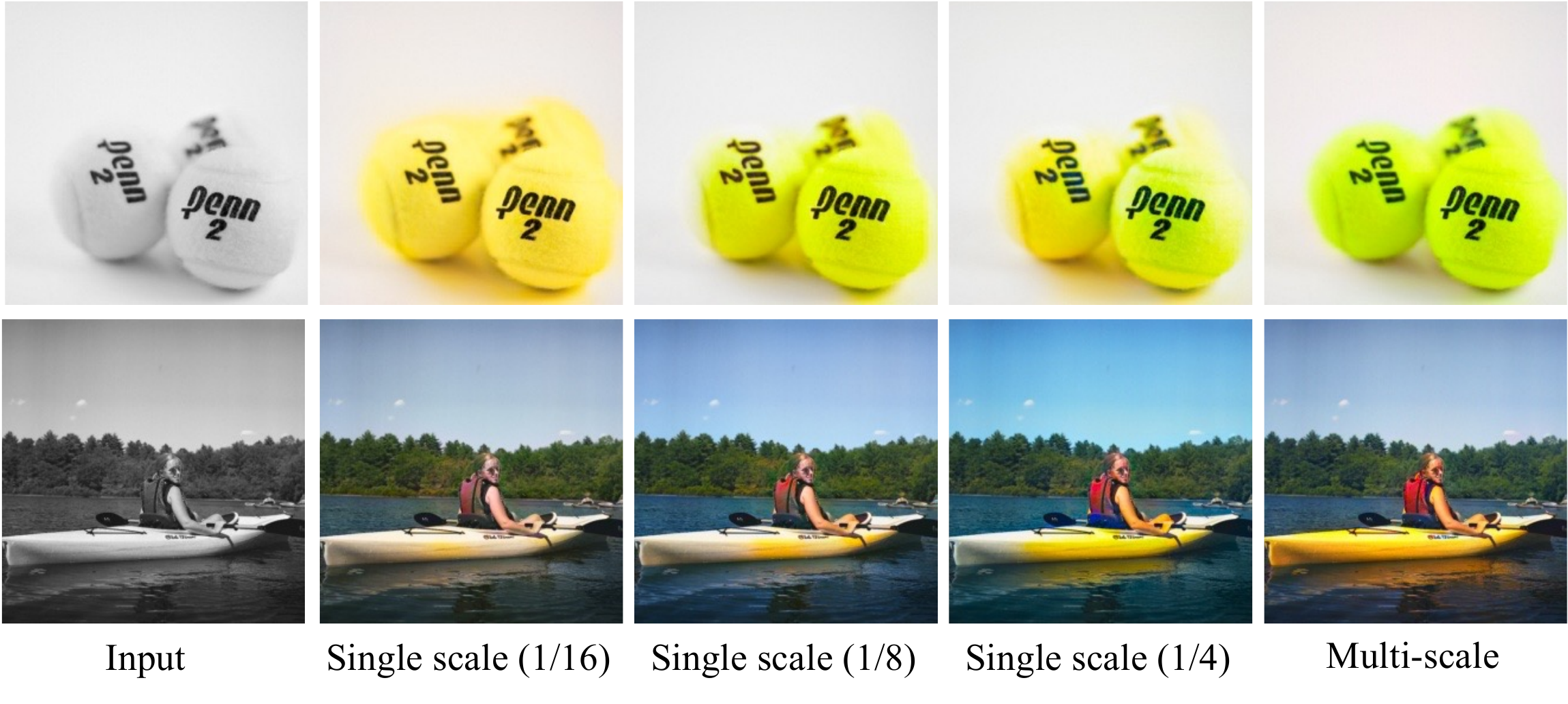}
  \caption{\textbf{Visual results} of ablation on feature scales.}
  \label{fig:ablation-scale}
\end{figure}

\noindent\textbf{Multi-scale vs. single scale.}
To evaluate the effect of features scale, we conduct 3 variants that use single scale features. 
The results in Figure~\ref{fig:ablation-scale} show that the 3 variants tend to produce visible color bleeding, inaccurate colorization results at the edges of objects (such as tennis in row 1), and semantically inconsistent colors for objects with different scales (such as people and kayak in row 2).
With multi-scale features, our full model captures more accurate identification feature of semantic boundaries and produces more natural and accurate colorization results. 

\noindent\textbf{Color decoder architecture.}
The color decoder architecture is designed for the purpose of using visual semantic information for learning color embeddings. We conduct ablation studies to validate the importance of each key component by modifying their arrangement. As shown in Table~\ref{tab:ablation_decoder}, we can see that both cross-attention layer and self-attention layer are essential for robust image colorization. This is because using merely the self-attention layer or the cross-attention layer lead to poor colorization results. Additionally, the sequence of self-attention and cross-attention layers also matters.

\begin{figure}[t]
  \centering
  \includegraphics[width=1.0\linewidth]{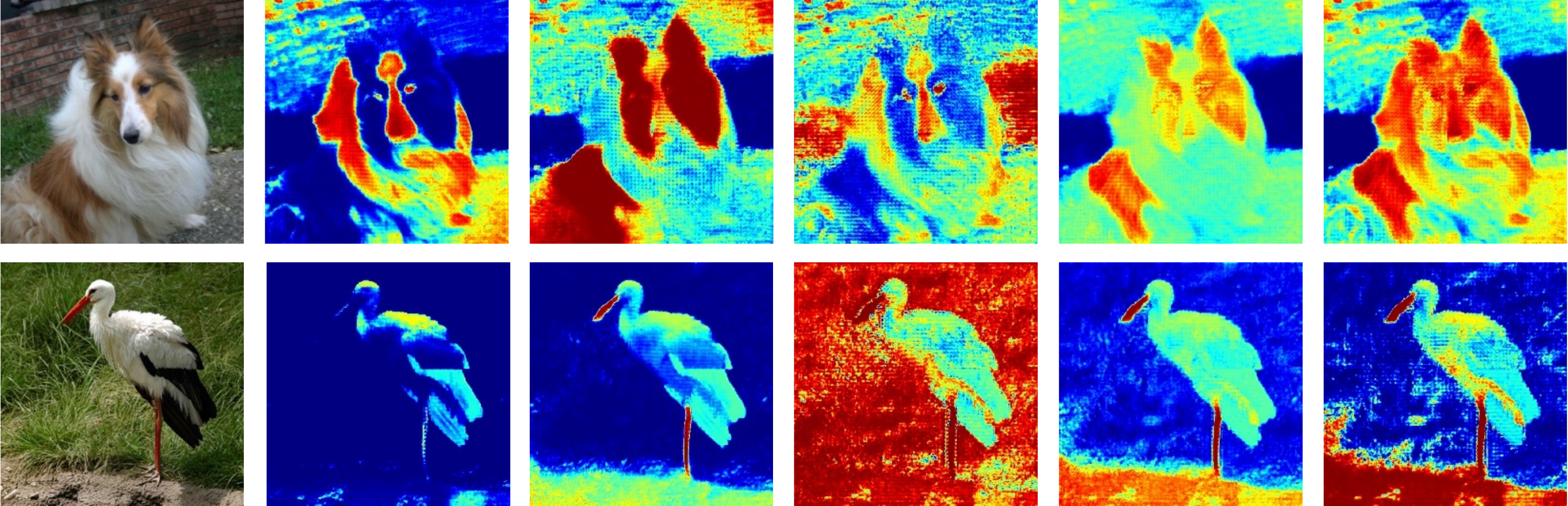}
  \caption{\textbf{Visualization of learned color queries.} The left column is the colorized image by our method, and columns on the right are the visualization of color queries. Red (blue) represents high (low) activation values.}
  \label{fig:query-vis}
\end{figure}

\begin{table}[t]
\small% \footnotesize
\centering
\setlength{\tabcolsep}{0.9\tabcolsep}
\begin{tabular}{cccc}
  \toprule
  \# of queries & FID$\downarrow$ & CF$\uparrow$ & $\Delta$CF$\downarrow$ \\
  \midrule
  20 & 4.02 & 37.86 & 0.35 \\
  50 & 3.96 & 38.00 & 0.21 \\
  100 & \textbf{3.92} & \textbf{38.26} & \textbf{0.05} \\
  200 & 3.96 & 37.71 & 0.50 \\
  500 & 3.93 & 37.88 & 0.33 \\
  \bottomrule
\end{tabular}
\caption{\textbf{Ablation on the number of color queries.} The model with 100 queries performs best on ImageNet dataset.}
\label{tab:ablation_query}
\end{table}

\noindent\textbf{Number of color queries.}
We vary the number of color queries to evaluate its effect. As shown in Table~\ref{tab:ablation_query}, the performance reaches to the peak at 100 queries, and no longer improves when the number of queries continues to increase. 
Our final model employs 100 queries, as this setting achieves optimal performance without excessive redundancy.
Several previous approaches \cite{zhang2016colorful, Weng2022CT2} consider colorization as a classification problem and quantify the $AB$ color space into 313 categories. In our approach, the learned color queries are adequate to represent color embeddings in the color space with much fewer categories. Interestingly, even with only 20 queries, our method outperforms the previous classification-based methods \cite{zhang2016colorful, Weng2022CT2} on FID.

\subsection{Visualizing the Color Queries}
\label{sec:visualize}
We visualize the learned color queries to reveal how it works. See Figure~\ref{fig:query-vis}.
Visualization results are obtained by sigmoiding the dot product of the single color query and the image feature map. 
It can be observed that each query specializes in certain feature regions, thus capturing semantically relevant color clues.
Taken the first row as an example, the first query attends on the forehead, nose and body of the dog, which may captures the white color embedding. The second and third queries focus on the fur of the dog and grass background regions, which may capture brown and green color embedding, respectively.

\subsection{Results on Real-world Black-and-white Photos}
\label{sec:real-result}
\begin{figure}[t]
  \centering
  \includegraphics[width=1.0\linewidth]{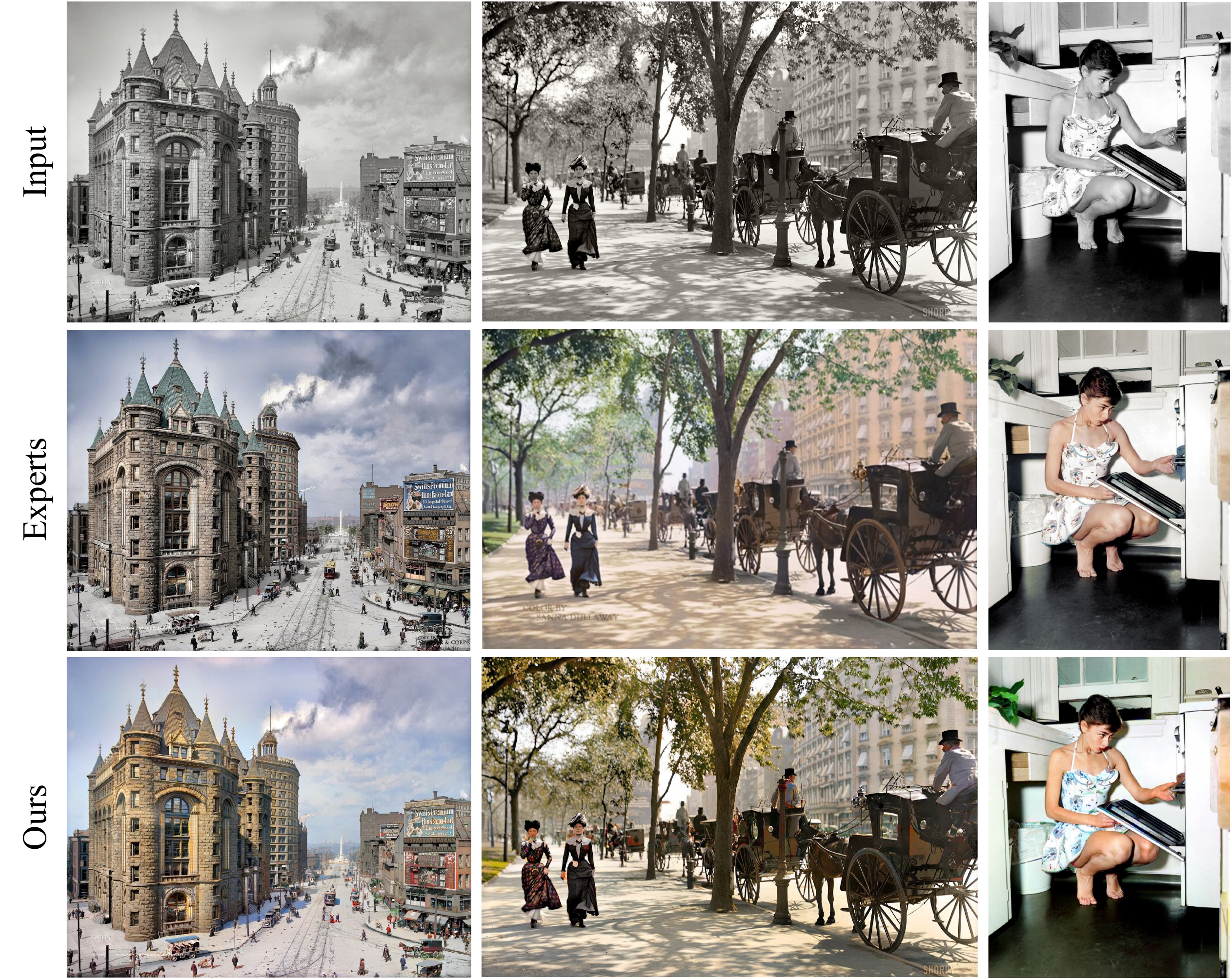}
  \caption{\textbf{Colorizing legacy photographs.} From top to bottom are respectively the input, the manually colorized results by human experts and our results.}
  \label{fig:colorize_real}
\end{figure}

We collect some real historical black-and-white photos to demonstrate the capability of our method in real-world scenarios. Figure~\ref{fig:colorize_real} shows the results of our method, as well as the manual colorization results by human experts\footnote{\href{https://www.reddit.com/r/ArchitecturePorn/comments/t55f2x/colorized_erie_county_savings_bank_niagara_street/}{reddit.com/r/ArchitecturePorn}}\footnote{\href{https://justsomething.co/34-incredible-colorized-photos/}{justsomething.co/34-incredible-colorized-photos}}, indicating the practicability of our approach.

\subsection{Limitation}
\label{sec:limits}

\begin{figure}[t]
  \centering
  \includegraphics[width=1.0\linewidth]{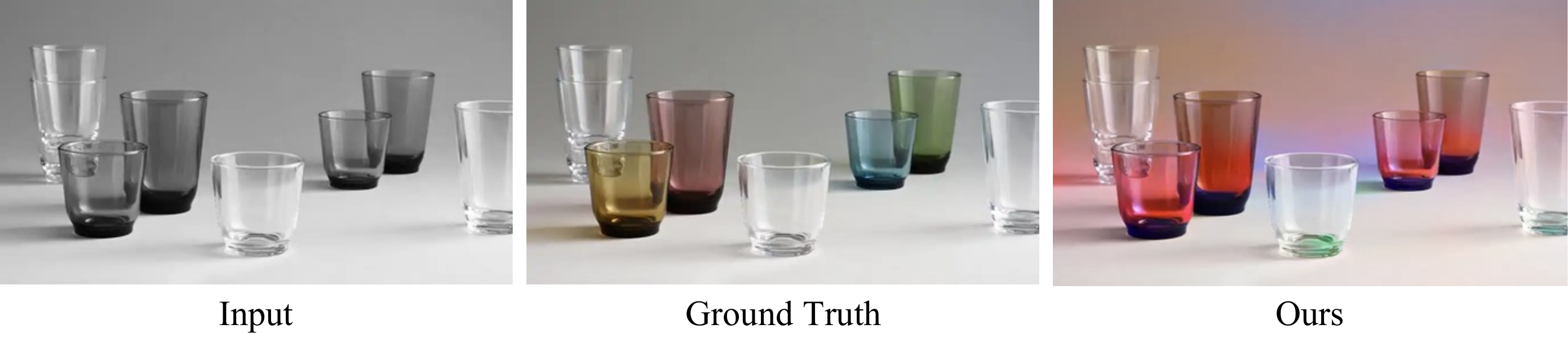}
  \caption{\textbf{Failure Cases.} Our method may still produce visual artifacts when coloring transparent/translucent objects.}
  \label{fig:fail}
\end{figure}

As shown in Figure~\ref{fig:fail}, there are still failure cases when dealing with images with transparent/translucent objects. Further improvement may require extra semantic supervision to help the network better understand such complex scenarios. Also, like most automatic colorization methods, our approach lacks user controls or guidance over the colors produced. Incorporating more user inputs such as text prompts, color graffiti in the colorization process will be a future work.

\section{Conclusion}
\label{sec:concl}
In this work, we propose an end-to-end method, called DDColor, for image colorization. The main contribution of DDColor lies in the design of two decoders: the color decoder, which learns semantic-aware color queries by utilizing query-based transformers, and the pixel decoder, which produces multi-scale visual features to optimize the color queries. Our approach surpasses previous methods in both performance and the ability to generate realistic and semantically consistent colorization.

{\small
\bibliographystyle{ieee_fullname}
\bibliography{egbib}
}

\appendix
% --- PDF will be split by an editor (e.g. macOS preview), so need to restart from page 1
% \setcounter{page}{1}

% --- repeat the title (AT: haven't found a more elegant way to do this...)
\twocolumn[
\centering
\Large
\textbf{DDColor: Towards Photo-Realistic and Semantic-Aware Image Colorization via Dual Decoders} \\
\vspace{0.5em}Appendix \\
\vspace{1.0em}
] %< twocolumn

\appendix

%%%%%%%%% BODY TEXT

In this supplementary document, we provide the following materials to complement the main manuscript:

\begin{itemize}[itemsep=0pt,topsep=0pt,parsep=0pt]
    \item Detailed network architecture of \methodname{};
    \item Additional qualitative results;
    \item Additional ablation study and visual results;
    \item Runtime analysis;
    \item More results on legacy black and white photos. 
\end{itemize}

\section{Detailed Network Architecture}
\label{sec:arch}

We list the detailed architecture of \methodname{} with a ConvNeXt-T\cite{liu2022convnet} backbone in Table~\ref{tab:arch}. The resolution of the input image is 256 $\times$ 256.

\begin{table*}[t]
\centering
\addtolength{\tabcolsep}{-2pt}
\vspace{4ex}
\begin{tabular}{c|c|c}
& Output size & \methodname{} \\
\hline

Stage 1&  64$\times$64$\times$96  &  
 \makecell[c]{Conv. 4$\times$4, 96, stide 4 \\ $\begin{bmatrix}\text{Depthwise Conv. 7$\times$7, 96}\\\text{Conv. 1$\times$1, 384}\\\text{Conv. 1$\times$1, 96}\end{bmatrix}$ $\times$ 3}  \\
\hline
Stage 2& 32$\times$32$\times$192  &  
 \makecell[c]{$\begin{bmatrix}\text{Depthwise Conv. 7$\times$7, 192}\\\text{Conv. 1$\times$1, 768}\\\text{Conv. 1$\times$1, 192}\end{bmatrix}$ $\times$ 3}  \\
\hline
Stage 3&  16$\times$16$\times$384  &  
 \makecell[c]{$\begin{bmatrix}\text{Depthwise Conv. 7$\times$7, 384}\\\text{Conv. 1$\times$1, 1536}\\\text{Conv. 1$\times$1, 384}\end{bmatrix}$ $\times$ 9}  \\
\hline
Stage 4& 8$\times$8$\times$768  &  
 \makecell[c]{$\begin{bmatrix}\text{Depthwise Conv. 7$\times$7, 768}\\\text{Conv. 1$\times$1, 3072}\\\text{Conv. 1$\times$1, 768}\end{bmatrix}$ $\times$ 3}  \\
\hline
Stage 5&  16$\times$16$\times$512  &  
 \makecell{PixelShuffle, scale 2 \\ Concat feat. from Stage 3 \\ Conv. 3$\times$3, 512 }\\
\hline
Stage 6&  32$\times$32$\times$512  &  
\makecell{PixelShuffle, scale 2 \\ Concat feat. from Stage 2 \\ Conv. 3$\times$3, 512 }  \\
\hline
Stage 7&  64$\times$64$\times$256  &  
 \makecell{PixelShuffle, scale 2 \\ Concat feat. from Stage 1 \\ Conv. 3$\times$3, 256 }  \\
\hline
{Stage 8} & 
{256$\times$256$\times$256 } & 
{PixelShuffle, scale 4}  \\
\hline
Color Dec. & 
 256$\times$100 & 
 \makecell[c]{
Conv. 1$\times$1, 256 feat. from Stage 5 \\ 
Conv. 1$\times$1, 256 feat. from Stage 6 \\ 
Conv. 1$\times$1, 256 feat. from Stage 7 \\ 
$\begin{bmatrix} \text{Conv. 1$\times$1, 256$\times$3} \\ \text{Cross-attn.} \\ \text{Self-attn.}\\ \text{Conv. 1$\times$1, 2048} \\ \text{Conv. 1$\times$1, 256}\end{bmatrix}$ $\times$ 9}    \\
\hline
Stage 9&  256$\times$256$\times$100  &  
\makecell{Dot Product feat. from Stage 8 \\ \& feat. from Color Dec. } \\
\hline
Stage 10&  256$\times$256$\times$2  &  
\makecell{Concat input \\ Conv. 1$\times$1, 2 }  \\
\hline
\end{tabular}
\vspace{3ex}
\normalsize
\caption{\textbf{Detailed architecture of \methodname{}}.}
\label{tab:arch}
\vspace{2ex}
\end{table*}

\section{Additional Qualitative Results}
\label{sec:add-qual}
\begin{figure*}[t]
  \centering
  \includegraphics[width=1.0\linewidth]{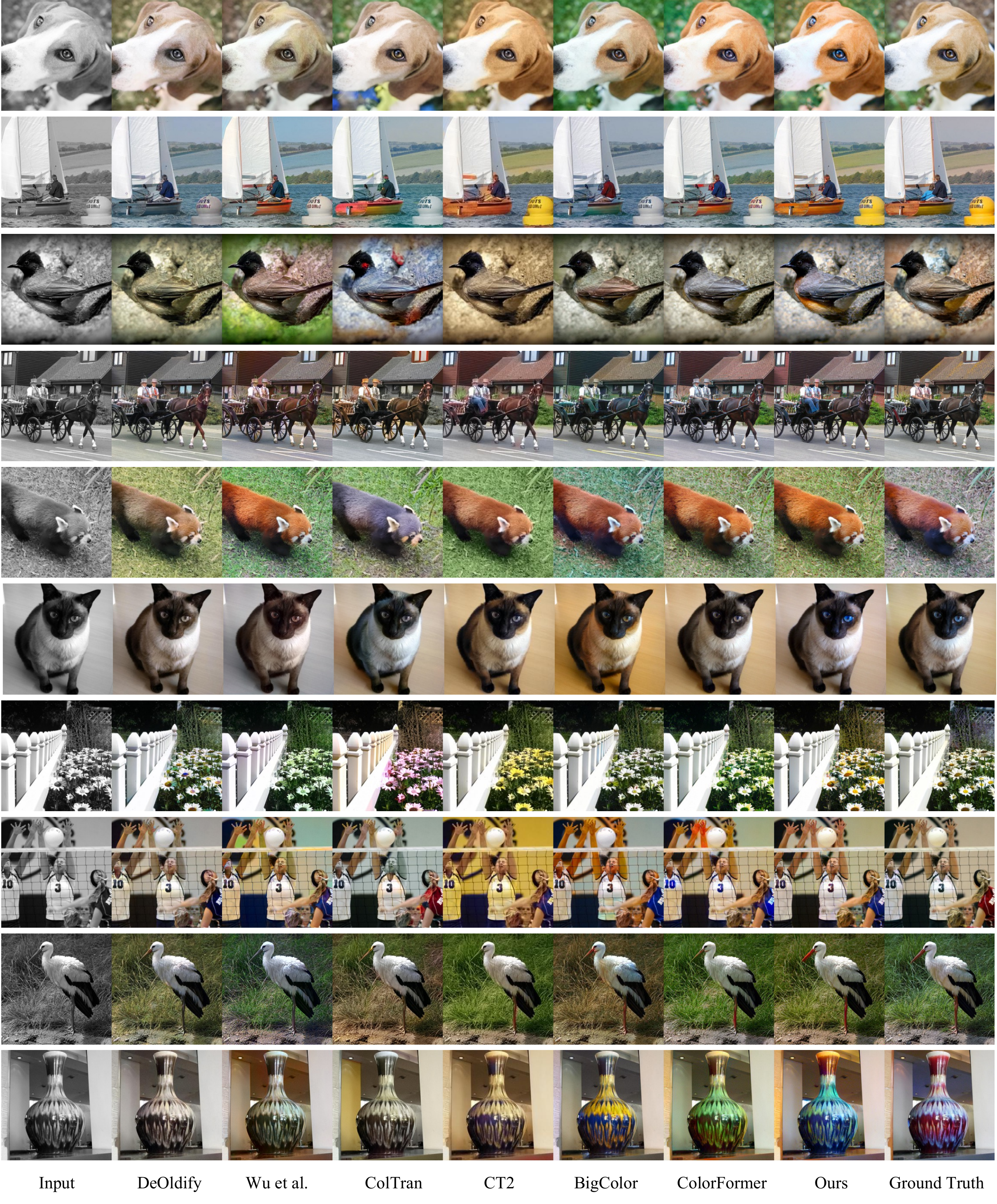}

  \caption{\textbf{More qualitative comparisons with previous colorization methods on ImageNet.}}
  \label{fig:imagenet-more}
\end{figure*}

\begin{figure*}[t]
  \centering
  \includegraphics[width=1.0\linewidth]{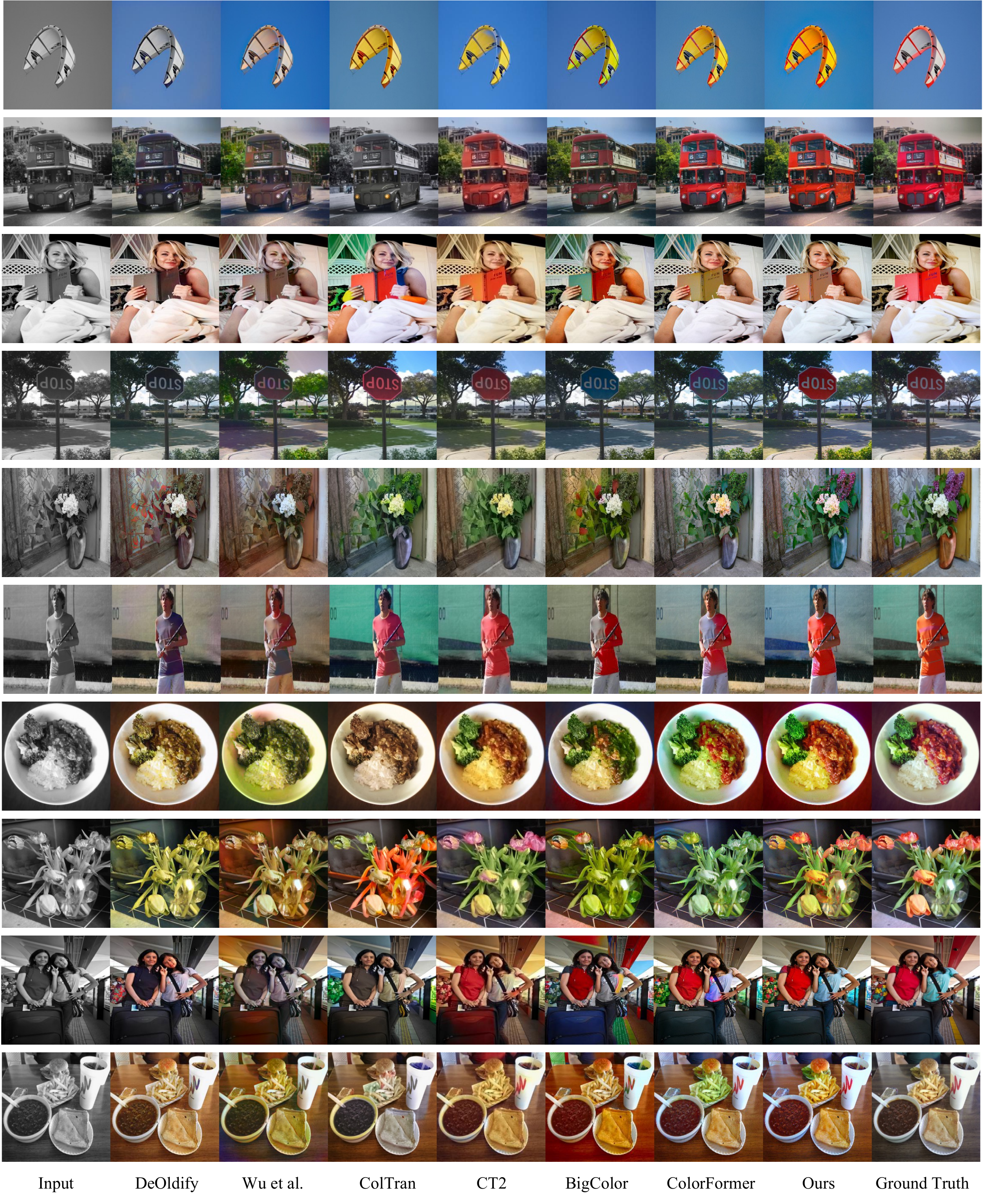}

  \caption{\textbf{Qualitative comparisons with previous colorization methods on COCO-Stuff.}}
  \label{fig:coco}
\end{figure*}

\begin{figure*}[t]
  \centering
  \includegraphics[width=1.0\linewidth]{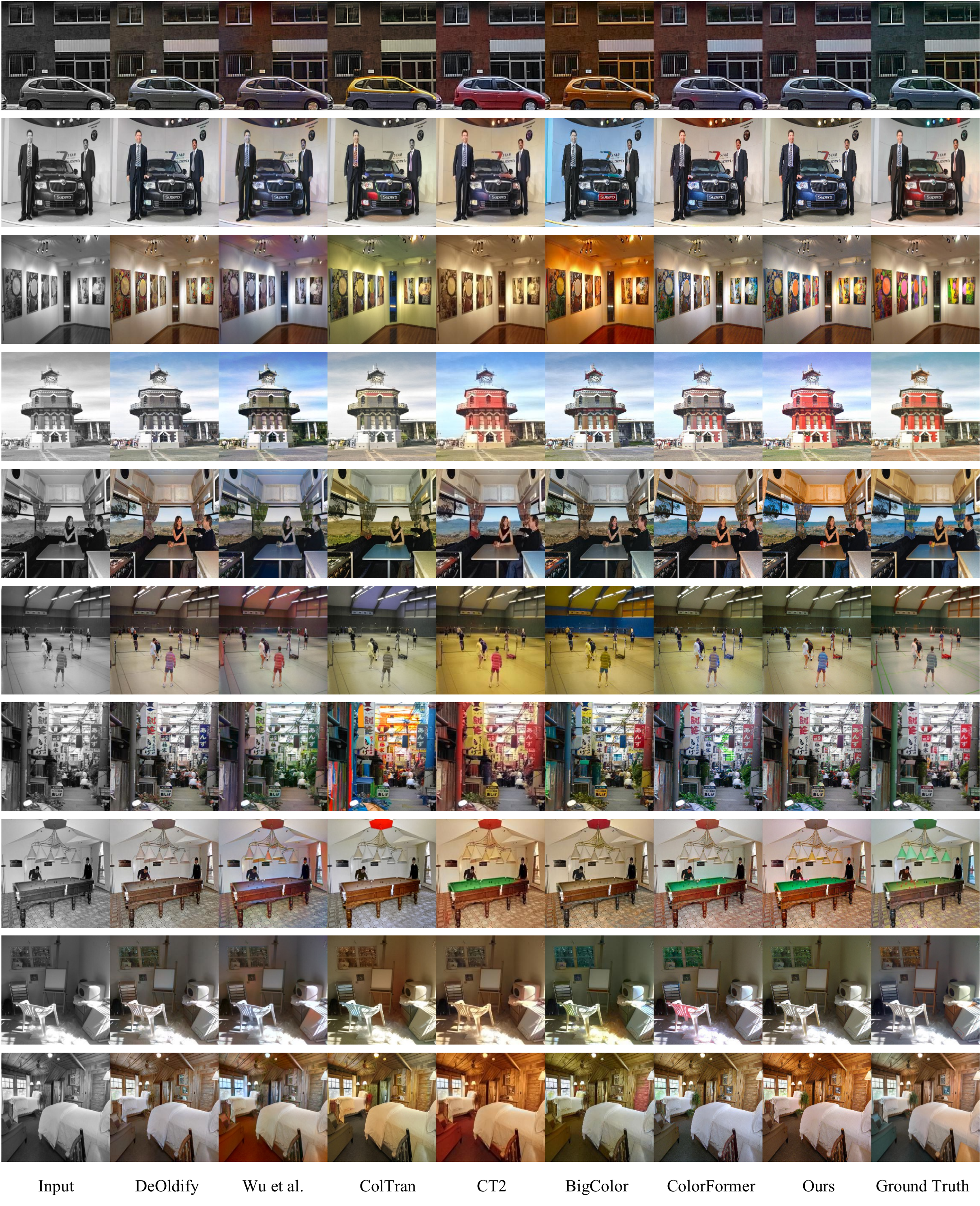}

  \caption{\textbf{Qualitative comparisons with previous colorization methods on ADE20K.}}
  \label{fig:ade20k}
\end{figure*}

Here, we show more qualitative comparisons with previous methods on ImageNet\cite{russakovsky2015imagenet} validation in Figure~\ref{fig:imagenet-more}. As in the main paper, we compare our method with DeOldify \cite{deoldify}, Wu \etal \cite{wu2021towards}, ColTran \cite{kumar2021colorization}, CT2 \cite{Weng2022CT2}, BigColor \cite{Kim2022BigColor} and ColorFormer \cite{Ji2022ColorFormer}.
The visual comparisons on COCO-Stuff\cite{caesar2018coco} and ADE20K\cite{zhou2017scene} are also presented in Figure~\ref{fig:coco} and \ref{fig:ade20k}, respectively.
It can be seen that our method achieves more natural and vivid results in diverse scenarios, and produces more semantically consistent colors for a variety of objects.

\section{Additional Ablation Study and Visual Results}
\label{sec:add-abla}
% \noindent\textbf{Backbone variants.}
We build four variants of our model with different ConvNeXt\cite{liu2022convnet} backbones, as detailed in Table~\ref{tab:ablation_backbone}. As can be seen, the backbone plays a key role in image colorization. We choice ConvNeXt-L due to its superior performance.

\begin{figure}[h]
  \centering
  \includegraphics[width=1.0\linewidth]{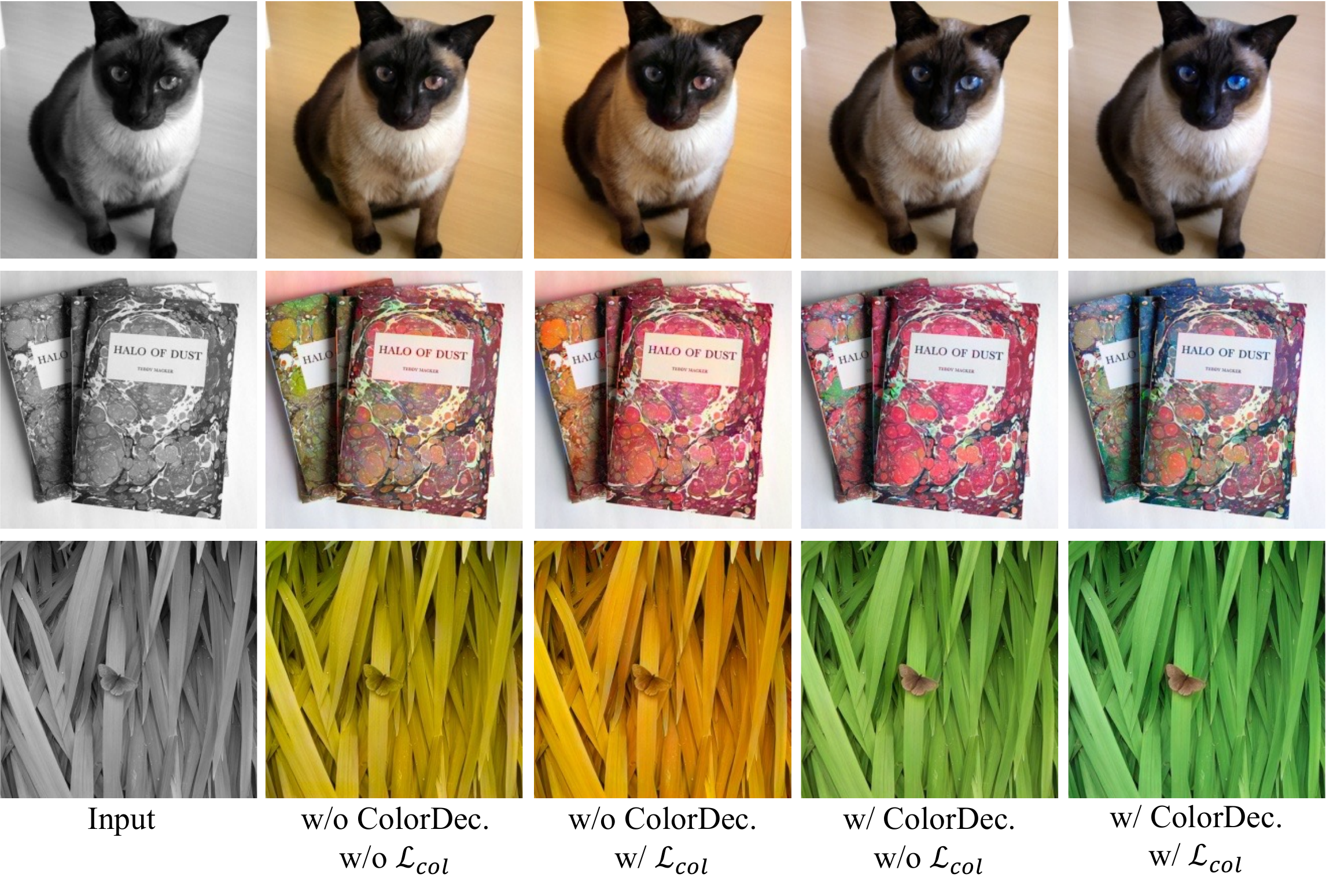}

  \caption{\textbf{More visual results of ablation on color decoder and colorfulness loss.}}
  \label{fig:ablation_module_supp}
\end{figure}

\begin{figure}[h]
  \centering
  \includegraphics[width=1.0\linewidth]{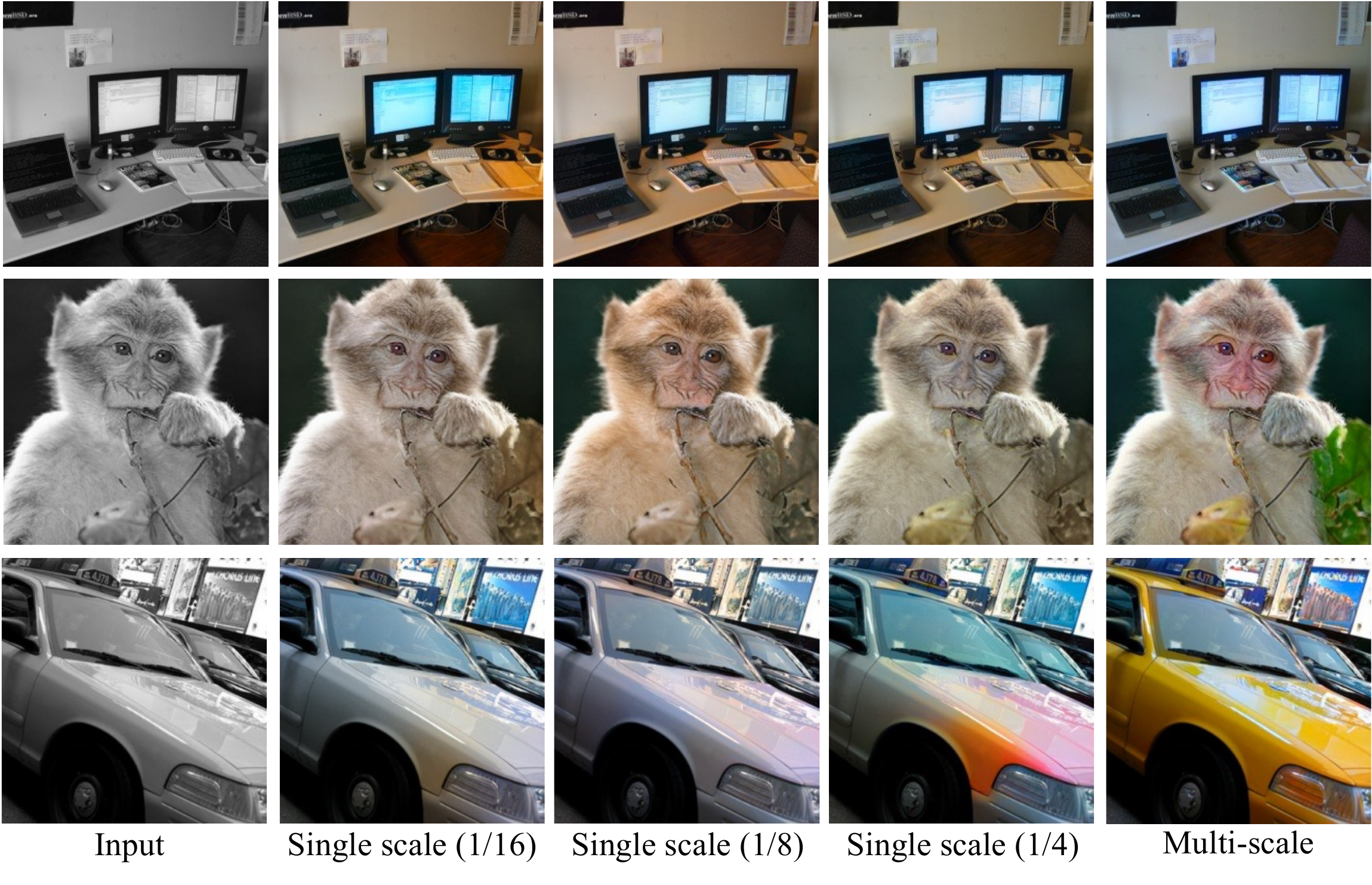}

  \caption{\textbf{More visual results of ablation on different feature scales.}}
  \label{fig:ablation_scale_supp}
\end{figure}

\begin{table*}[t]
\centering
\setlength{\tabcolsep}{0.7\tabcolsep}
\begin{tabular}{cccccr}
  \toprule
  Model Name & Backbone & FID$\downarrow$ & CF$\uparrow$ & $\Delta$CF$\downarrow$ & Params \\
  \midrule
  DDColor-T & ConvNeXt-T & 4.38 & 37.66 & 0.55 & 55.0M\\
  DDColor-S & ConvNeXt-S & 4.25 & 38.10 & 0.11 & 76.6M\\
  DDColor-B & ConvNeXt-B & 4.06 & 38.15 & 0.06 & 116.2M\\
  DDColor-L & ConvNeXt-L & \textbf{3.92} & \textbf{38.26} & \textbf{0.05} & 227.9M \\
  \bottomrule
\end{tabular}
\vspace{4ex}
\caption{\textbf{Backbone variants.} We build four variants of our DDColor based on backbones of different sizes. The overall performance improves with the increase of the scale of the backbone network.}
\label{tab:ablation_backbone}
\end{table*}

More visual results on ablations of color decoder, colorfulness loss, and different visual feature scales are shown in Figure~\ref{fig:ablation_module_supp} and Figure~\ref{fig:ablation_scale_supp}.

\section{Runtime Analysis}
\label{sec:runtime}
Our method colorizes grayscale images of resolution 256$\times$256 at 25 FPS / 21 FPS using ConvNeXt-T / ConvNeXt-L as the backbone. The inference speed of our end-to-end method is $\times 96$ faster than the previous transformer-based method \cite{kumar2021colorization}.
% The proposed \methodname{}  is an end-to-end network. With ConvNeXt-T\cite{liu2022convnet} as the backbone, our algorithm colorizes images of 256$\times$256 at 25 FPS with 55.0M model parameters. For the larger version with the ConvNeXt-L\cite{liu2022convnet} backbone, our algorithm colorizes images at 22 FPS with 227.9M model parameters. 
All tests are performed on a machine with an NVIDIA Tesla V100 GPU.

\section{More Results on Legacy Black and White Photos}
\label{sec:app}
\begin{figure*}[t]
  \centering
  \includegraphics[width=0.68\linewidth]{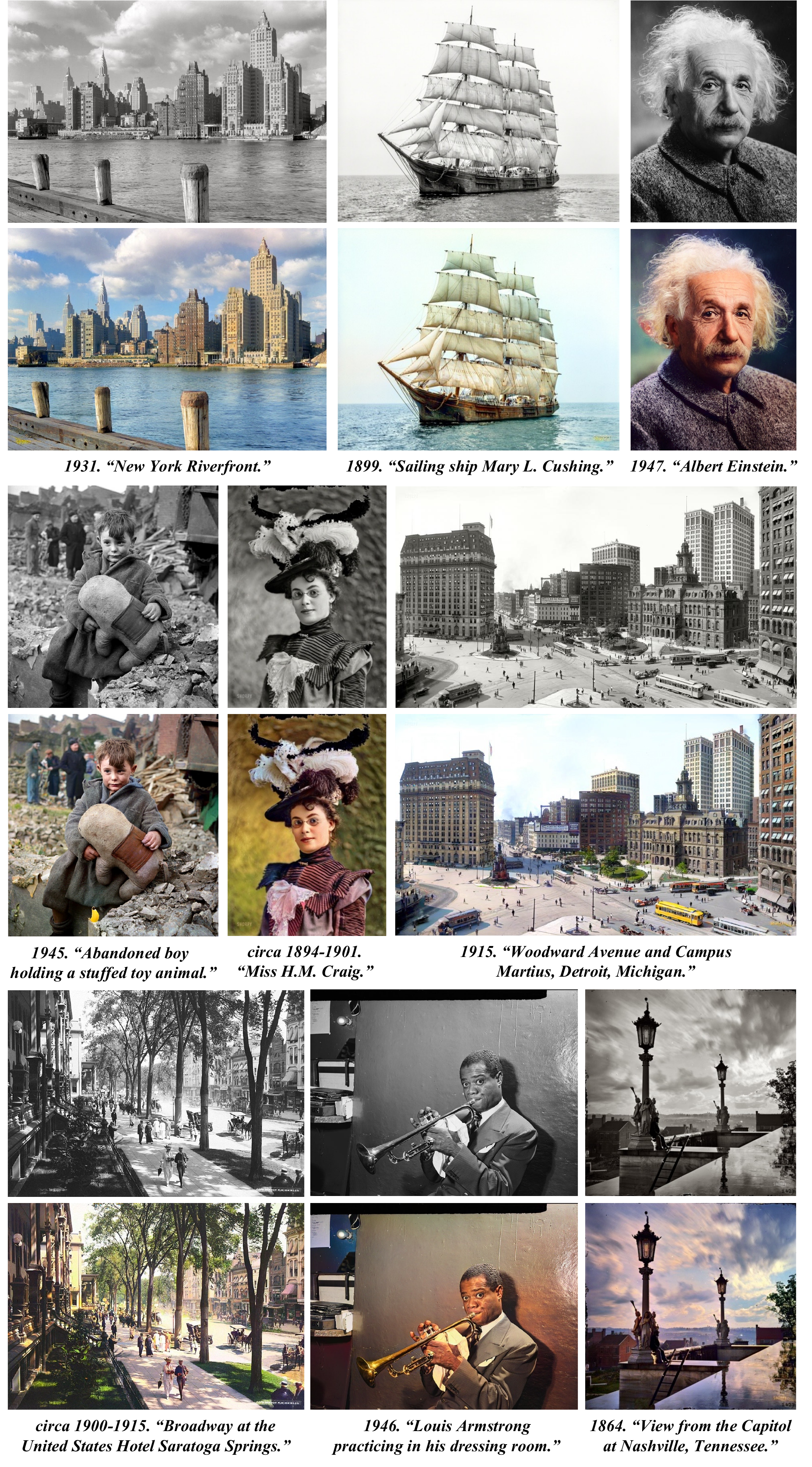}

  \caption{\textbf{More results on legacy black and white photos.}}
  \label{fig:supp-real}
\end{figure*}

More colorization results on legacy black and white photos are shown in Figure~\ref{fig:supp-real}, demonstrating the generalization capability of our method.

% \begin{figure}[t]
%   \centering
%   \includegraphics[width=0.7\linewidth]{figures/ablation_module_supp.pdf}

%   \caption{\textbf{More visual results of ablation on color decoder and colorfulness loss.}}
%   \label{fig:ablation_module_supp}
% \end{figure}

% \begin{figure}[t]
%   \centering
%   \includegraphics[width=0.7\linewidth]{figures/ablation_scale_supp.pdf}

%   \caption{\textbf{More visual results of ablation on different feature scales.}}
%   \label{fig:ablation_scale_supp}
% \end{figure}

% \clearpage

% {\small
% \bibliographystyle{ieee_fullname}
% \bibliography{egbib}
% }

% \end{document}

\end{document}